# Extracting Problem and Method Sentence from Scientific Papers: A Context-enhanced Transformer Using Formulaic Expression Desensitization

**Yingyi Zhang [a], Chengzhi Zhang [b, *]**

[a] Department of Archives and E-government, Soochow University, Suzhou, China, yyzhang9@suda.edu.cn

[b] Department of Information Management, Nanjing University of Science and Technology, Nanjing, China, zhangcz@njust.edu.cn

**Abstract**: Billions of scientific papers lead to the need to identify essential parts from the massive text. Scientific research is an activity from putting forward problems to using methods. To learn the main idea from scientific papers, we focus on extracting problem and method sentences. Annotating sentences within scientific papers is labor-intensive, resulting in small-scale datasets that limit the amount of information models can learn. This limited information leads models to rely heavily on specific forms, which in turn reduces their generalization capabilities. This paper addresses the problems caused by small-scale datasets from three perspectives: increasing dataset scale, reducing dependence on specific forms, and enriching the information within sentences. To implement the first two ideas, we introduce the concept of formulaic expression (FE) desensitization and propose FE desensitization-based data augmenters to generate synthetic data and reduce models' reliance on FEs. For the third idea, we propose a context-enhanced transformer that utilizes context to measure the importance of words in target sentences and to reduce noise in the context. Furthermore, this paper conducts experiments using large language model (LLM) based in-context learning (ICL) methods. Quantitative and qualitative experiments demonstrate that our proposed models achieve a higher macro $F_1$ score compared to the baseline models on two scientific paper datasets, with improvements of 3.71% and 2.67%, respectively. The LLM based ICL methods are found to be not suitable for the task of problem and method extraction.

## 1. Introduction

The number of scientific papers in the world has reached a billion, and a previous study has indicated that this number continues to grow at a rate of 3% annually (Bornmann & Mutz, 2015). Massive scientific papers surpasses human processing and comprehension capabilities. Scientific research is an activity from putting forward problems to using methods to solve them (Wilson, 1952). Problems and methods are two essential components of scientific papers (Heffernan & Teufel, 2018; Luo et al., 2022). The problem is the subject of the scientific paper and the technical problem, while the method is the process researchers used to achieve the goal and the details of the experimental setup. Problems and methods are described in problem and method sentences in scientific papers. By analyzing these sentences, we can rapidly grasp the meaning of problems and methods, as well as the main point of the scientific papers. Thus, this paper aims to extract problem and method sentences, which has various downstream applications, e.g., knowledge graph construction (Oelen et al., 2021), academic search (Boudin et al., 2010; Dernoncourt et al., 2016; Neves et al., 2019; Safder & Hassan, 2019), and scientific paper summarization (Dernoncourt et al., 2016; Teufel & Moens, 2002).

---

[*] **Corresponding author: Chengzhi Zhang.**

Recently, several studies have focused on extracting problem and method sentences from scientific papers, with the most relevant ones being Sakai and Hirokawa (2012) and Wang et al. (2020). Sakai and Hirokawa (2012) extracted problem sentences and manually constructed a dataset comprising 300 abstracts. Wang et al. (2020) extracted method sentences and manually created a testing dataset of 100 abstracts. In addition to these problem and method sentence datasets, there are other scientific paper sentence annotation datasets available, such as the NIC dataset (1000 abstracts), which was annotated with sentence categories including background (Kim et al., 2011), the DRI dataset (40 full texts), annotated with sentence categories such as background and results (Fisas et al., 2015), and the SciContri dataset (783 contribution paragraph), annotated with contribution statements (Chao et al., 2023). As evident from these examples, the labor-intensive nature of manual annotation leads to the creation of small-scale datasets. Previous research generally conducted experiments directly on these small-scale datasets (Mutlu et al., 2020; Tokala et al., 2023) . For deep neural networks, small-scale datasets fail to provide sufficient information. In this situation, neural networks are prone to memorizing particular forms that do not generalize well (Shorten et al., 2021). Overreliance on these forms diminishes the models' generalization ability and can degrade their performance.

Therefore, this paper aims to mitigate the overfitting issue caused by small-scale datasets in the task of problem and method sentence extraction from three perspectives: (a) expanding the size of the training dataset, (b) diminishing the models' dependence on specific patterns, and (c) enriching the information within sentences.

**Table 1. The example of reliance on formulaic expressions (FEs). The word in bold indicates the FE-related word within the sentence. "CITE" denotes the citation information within the sentence.**

| Category | Content | Prediction results |
|---|---|---|
| **Target Sentence** | More specifically, we combine a probabilistic topological field parser **for** German CITE with the HPSG parser of CITE. | **Human annotation:** Method sentence<br>**Predicted label by baselines**: Problem and method cooccurrence sentence |

To implement strategies (a) and (b), we utilize the concept data augmentation. Data augmentation involves creating synthetic data with minor changes (Shorten et al., 2021). These synthetic sentences not only increases the size of the original training dataset but also reduce the models' reliance on high-frequency patterns (Shakeel et al., 2020). Existing data augmentation methods employ various techniques for word selection, such as random selection (Kobayashi, 2018) and TFIDF-based strategies (Xie et al., 2020) , or they generate new sentences using pre-trained language models like GPT-3 (Ding et al., 2022). However, within these approaches, the random strategy involves randomly choosing words within a sentence for substitution, whereas the TF-IDF strategy concentrates on  replacing less significant words. Desptie these efforts, data augmentation methods using these word selection strategies often fail to alleviate the model's reliance on specific patterns. Likewise, sentences generated by LLMs may still retain particular patterns and may not effectively decrease the model's reliance on these specific patterns.

Unlike previous studies, and informed by observation of problem and method sentences, we introduce formulaic expression (FE) desensitization-based data augmenters. Formulaic expression, or FEs, are pre-stored word strings in human memory that are frequently employed to convey the intent and function of sentences(Wray, 2000). In problem and method sentences, FEs typically manifest as n-gram phrase that are commonly found in these sentences, such as “was used for”(Iwatsuki & Aizawa, 2021). When deep neural network encounter FEs, they tend to hastily predict sentences as problem and method sentences. As illustrated in Table 1, the word “for” in this sentence often appears in FEs, which are frequently used to indicate that a method is employed to adddress a specific problem. Consequently, the baseline model incorrectly predicts this sentence as describing both the problem and method. Clearly, the word “German” following “for” does not related to a problem. Therefore, this paper replaces FEs with other words to reduce models’ reliance on FE patterns. This process is termed FE desensitization. As mentioned above, we pose the first research question:

**RQ1:** Can the FE desensitization-based data augmenter alleviate models’ dependence on FE forms?

To implement (b), our goal is to integrate context into the models. To illustrate the role of context, we present an example in Table 2. In this case, the target sentence includes the method “Glosser” and the problem “support reading and learning to read in a foreign language”. Among the adjacent sentences, the following one contains the term “Glosser” and offers a detailed description of it. This observation demonstrates that context can enrich the semantics of the target sentence. However, it is important to note that adjacent sentences may also include unrelated information, such as the term "English." Consequently, directly using adjacent sentences could introduce noise into the model.

**Table 2. An example of the context mechanism. The words in blue and green denote the method and problem content within the target sentence, respectively. The bold words indicate valuable context content, whereas the words in yellow indicate noise within the context.**

| Sentence category | Sentence content |
|---|---|
| Target sentence | Glosser is designed to support reading and learning to read in a foreign language. |
| Preceding sentence | None |
| Following sentence | There are four language pairs currently **supported by Glosser**: English-Bulgarian, English-Estonian, English-Hungarian, and French-Dutch |

In this paper, we propose a model that efficiently integrates context with the target sentence while minimizing noise. Inspired by the work of  Yu et al. (2020)  and Yang et al. (2022), which emply an image as a query in a transformer to learn the importance of words in sentences (Vaswani et al., 2017), we introduce a context-enhanced transformer. This transformer uses the context as the query to identify the significant content within the target sentence, thereby enriching the representation of the sentence’s semantics. This approach aims to preserve the full semantic content of the target sentences. As mentioned above, we pose the second research question:

**RQ2:** Can the context-enhanced transformer effectively incorporate context into the original sentence while reducing noise?

This paper manually constructs two datasets, i.e., the abstract dataset and the full-text dataset. For both datasets, the definitions of problem and method sentences are derived from prior research. Additionally, through dataset analysis, we introduce a new category of sentences, referred to as problem and method co-occurrence sentences (Abbr. co-occurrence sentences). Recognizing that annotators may make errors, we employ a cross-weight-based label error detection and correction framework (Z. Wang et al., 2019). With these two datasets, we pose the third research question:

**RQ3:** Do the models presented in this paper achieve a higher macro $F_1$ score than the baselines in problem and method sentence extraction tasks, based on small-scale abstract and full-text datasets?

Nowadays, large language models (LLMs) have been developed and have shown promising performance in generation and annotation tasks through in-context learning (ICL), without the need for model fine-tuning (Deng et al., 2022; Madaan et al., 2023). In-context learning leverages prompt information to yield results from LLMs, either with or without a small amount of demonstration. Consequently, it is well-suited for low-resource tasks. However, there is a lack of research exploring the application of LLMs in the task of extracting sentences from scientific papers and evaluating their performance. Thus, this paper conducts experiments on the two datasets to evaluate the performance of LLM-based ICL. As mentioned above, we pose the fourth research question.

**RQ4:** Can LLM-based ICL efficiently extract problem and method sentences from scientific papers?

The contribution of this study lies in four aspects. First, we introduce FE desensitization-based data augmenters to increase the scale of the training dataset and reduce models' dependence on FE forms. Second, we design a context-enhanced transformer to integrate context with the target sentence while reducing the noise in the context. Third, we define three categories of sentences, i.e., problem sentence, method sentence, and co-occurrence sentence, and manually construct two datasets based on these definitions. Experiments conducted on these datasets demonstrate that our proposed models yield higher macro F1 score than baselines. Fourth, we evaluate the performance of LLM-based ICL on the task of sentence extraction from scientific papers.

The data sample and source code of this paper are freely available at the GitHub website: https://github.com/YingyiZhang/sentence-extraction-from-scientific-paper

# 2. Literature Review

In this section, we first analyze the methods used for sentence extraction from scientific papers. Then, we sort out the data augmenters used in the sentence extraction task. Since the problem and method sentences should be defined before annotation, this section also delineates the definitions of problem and method sentences.

### 2.1 The method for sentence extraction from scientific papers

In sentence extraction tasks, abstracts are frequently chosen as the primary source material. Various neural network models have been applied to this task, including CNN (R. Wang et al., 2020), LSTM (Mutlu et al., 2020), BiGRU (Gonçalves et al., 2020), and SCIBERT (Beltagy et al., 2019). Furthermore, some studies have extended their efforts to extract sentences from the full text of scientific papers. In full text sentence extraction, traditional statistical methods are commonly employed and complemented with features, such as IMRD structure (Kovacevic et al., 2012) and citation information (Fisas et al., 2015). Unlike these approaches, Quatra & Cagliero (2022) used BERT-based models. In their approach, they first delineate the paragraph structure and subsequently append this definition to the target sentences. However, it is crucial to recognize that the terminology used to describe paragraph structure can vary across different scientific papers, necessitating a standardized approach. Moreover, the quality of the paragraph structure definition can significantly influence the model's performance.

Few researchers realize the significance of context in the sentence extraction task (Dernoncourt et al., 2016; Jin & Szolovits, 2018; Tokala et al., 2023; Yang et al., 2022). In these studies, context was taken into account within sequential models, where all sentences in the abstract are treated as a sequence and input into an RNN-based module. These models have two key characteristics. First, the predicted result of adjacent sentences can influence the prediction of the current sentence. Therefore, these models are well-suited for datasets with dense and sequential labels, such as abstract datasets with objectives, backgrounds, methods, and results labels. However, as problem and method sentence labels are sparse in full text, sequential models are not suitable for full-text sentence extraction. Second, the simultaneous input of all sentences into the model can lead to substantial memory consumption.

In this paper, we propose a context-enhanced transformer to integrate context with the target sentence and reduce the noise in the context. Unlike sequential models employed in previous studies (Dernoncourt et al., 2016; Jin & Szolovits, 2018; Tokala et al., 2023; Yang et al., 2022), our proposed model does not directly input all the context into the models, making it more suitable for datasets with sparse labels.

### 2.2 Data augmenters for sentence extraction task

Data augmenters for sentence extraction encompass both word-level and sentence-level methods. Among word-level augmenters, the random method is frequently employed (Wei & Zou, 2019). In this approach, the words to be replaced and the new words to be inserted are chosen randomly. However, random word replacement does not account for the significance of words within the sentence, and the random selection of new words can introduce noise. To account for word importance, TFIDF augmenters measure word significance and replace words with low TFIDF values (Xie et al., 2020). To minimize noise, embedding augmenters use embeddings such as GloVe (Pennington et al., 2014) to replace randomly selected words with their synonyms (W. Y. Wang & Yang, 2015). The contextual embedding augmenter takes a different approach by first randomly masking words and then using a pre-trained language model to predict the masked word (Wu et al., 2019).

Sentence-level augmenters often employ back translation methods(Graça et al., 2019; Maier Ferreira & Reali Costa, 2020). For instance, an English sentence can be translated into another language and then translated back into English. Another approach involves the use of generation models, such as large language models (LLMs) and generative adversarial network (GANs), to create new sentences (Ding et al., 2022; Zhou et al., 2020). For example, Ding et al. (2022) used Generative Pre-trained Transformer 3 (GPT-3) along with prompt information to generate entities and then generate sentences using these generated entities.

Previous augmenters selected words for replacement either randomly or using the TFIDF method (Wei & Zou, 2019; Wu et al., 2019; Xie et al., 2020), which has proven ineffective in effectively replacing FEs in sentences and reducing their prominence. The generation-based method also does not resolve the issue of FE dependency.

In this paper, we introduce FE desensitization-based data augmenters to desensitize FEs in sentences and thereby expand the training datasets.

### 2.3 Problem and method sentence definition

We first analyze the definition of the problem sentence. Previous studies have defined problem sentences from two perspectives (Sakai & Hirokawa, 2012), i.e., sentences that describe the technical problem and sentences that describe the paper's subject. The technical problem represents the gap between the methods used and the ideal solution. With regard to the second aspect, there are similar expressions used in previous studies. For instance, some studies have employed the terms "purpose" (Yamamoto & Takagi, 2005) and "aim" (Teufel & Moens, 2002) instead of "subject" to convey the same idea.

Then, we delve into the definition of method sentences. The method sentence can be characterized from various angles. First, it is defined as a sentence that describes the process employed by researchers to achieve a goal (Liakata et al., 2010; Liu et al., 2013; Yamamoto & Takagi, 2005). Second, some studies consider a method sentence to contain the details of the experimental setup, such as the data, models, framework, and metrics used in the experiment) (Fisas et al., 2015; Liakata et al., 2010). Third, few studies regard sentences containing information about a method mentioned or used in previous works as method sentences (Iwatsuki & Aizawa, 2021; Liakata et al., 2010).

In this paper, we adopt the definition from previous studies to define method sentences and problem sentences. In addition, we introduce a new category of sentences that has been overlooked by previous studies: the problem and method co-occurrence sentence.

## 3. Methodology

This section introduces FE desensitization-based data augmenters and then provides a detailed description of the context-enhanced transformer. To integrate these two methods, we train the context-enhanced transformer using augmented datasets. Furthermore, to analyze the perofmance of LLM on the problem and method sentence extraction task, this paper designs LLM-based ICL to extract problem and method sentences from both abstract and full-text datasets.

### 3.1 FE desensitization-based data augmenter

The FE desensitization-based data augmenter consists of two steps. First, we select and replace FEs. Second, we use a sentence quality discriminator to determine whether the augmented sentence should be added to the original training datasets. In FE selection, we have designed three distinct strategies: the dependency-tree-based strategy, the LLM-based strategy, and the open-source-based strategy. The dependency-tree-based strategy and LLM-based strategy select FEs from the original training datasets, whereas the open-source-based strategy use FEs provided by Iwatsuki and Aizawa (2021). For assessing the quality of the augmented sentences, we employ two strategies, i.e., the "true-discrimination" strategy and the "wrong-discrimination" strategy.

**(1) Dependency-tree-based FE selection strategy**

Inspired by the pattern selection method proposed by Dang & Aizawa(2008), we designed the dependency-tree-based FE selecting strategy, which involves three steps. An example is illustrated in Figure 1.

**Step1:** we utilize Stanford CoreNLP (Manning et al., 2014) to construct the dependency tree of the sentence.

**Step 2:** we identify problem and method phrases in the sentence and use the Stanford CoreNLP to generate the dependency subtree of these phrases. The recognition of problem and method phrases can be achieved through the use of existing phrase recognition models (Luan et al., 2017).

**Step 3:** we select FEs with the aid of the dependency tree and replace FEs with the symbol "_". Since FEs are words related to the problem and method content in the sentence, three types of words in the dependency tree are chosen as FEs, i.e., the root of the dependency subtree, the parent nodes of the dependency subtree, and the preposition connecting the parent node to the dependency subtree. We then construct a FE set and add the selected FEs to the set. In Figure 1, the four words "present", "method", "finding", and "for" are chosen as FEs.

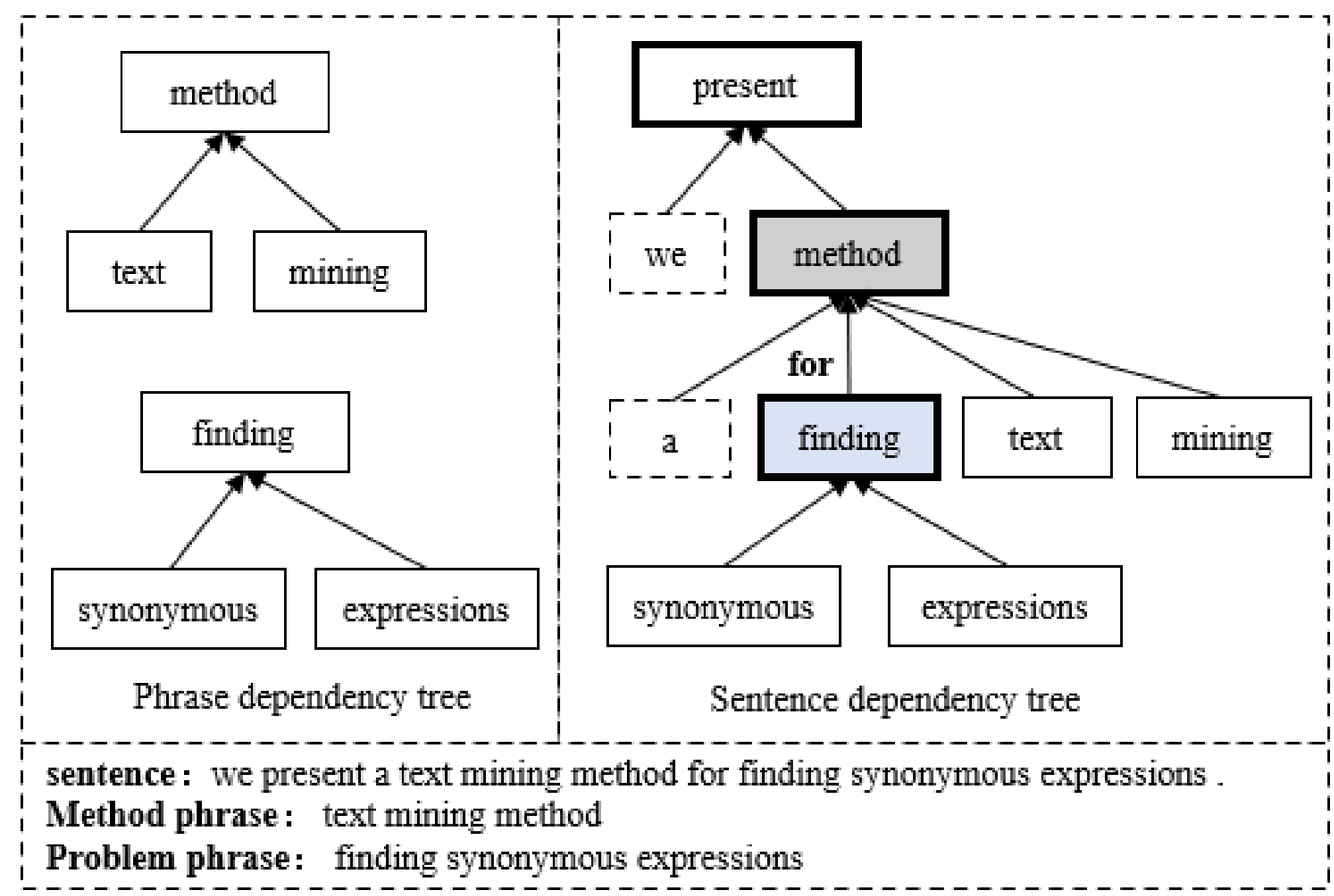

**Figure 1. The framework of the dependency-tree-based FE selection strategy.**

**(2) LLM-based FE selection strategy**

In the LLM based strategy, we design a prompt to use LLM to extract FEs from sentences. The prompt consists of three components, i.e., query question, a set of demonstration, and the target sentence. The query question is "Please extract formulaic expression from sentences I provide below." We provide four demonstrations, which are selected from the dataset provided by Iwatsuki and Aizawa (2021). The selected FEs in the sentence is then replaced with the symbol "_".

**(3) Open-source-based FE selection strategy**

We utilize the FEs extracted by Iwatsuki and Aizawa (2021), who annotated various categories of sentences and subsequently extracted FEs from these sentences. For instance, they annotated sentences related to "showing the main problem in the field" and "description of the process", which can be regarded as problem sentences and method sentences, respectively. From these sentences, they extracted 288 problem FEs and 1674 method FEs. After conducting stemming on these FEs, the present paper replaces these FEs with the symbol "_".

**(4) Sentence Quality Discrimination**

This paper employs two sentence quality discrimination methods, i.e., the "true-discrimination" strategy and the "wrong-discrimination" strategy. The "True-discrimination" strategy, as proposed by Zeng et al. (2020), is illustrated in Figure 2. In this strategy, the augmented sentences are input into a quality discriminator, which is a sentence extraction model trained on the original training dataset. The discriminator is used to predict the category of the augmented sentences. If the predicted category matches the category of the original sentence, the augmented sentence is added to the original training dataset. However, augmented sentences that are correctly predicted may be considered simple cases and not particularly useful for the current models. On the other hand, augmented sentences that are wrongly predicted may contain valuable information for the discriminators. Thus, this paper also employs the "wrong-discrimination" strategy, which selects the wrongly predicted augmented sentence and add them to the original training dataset.

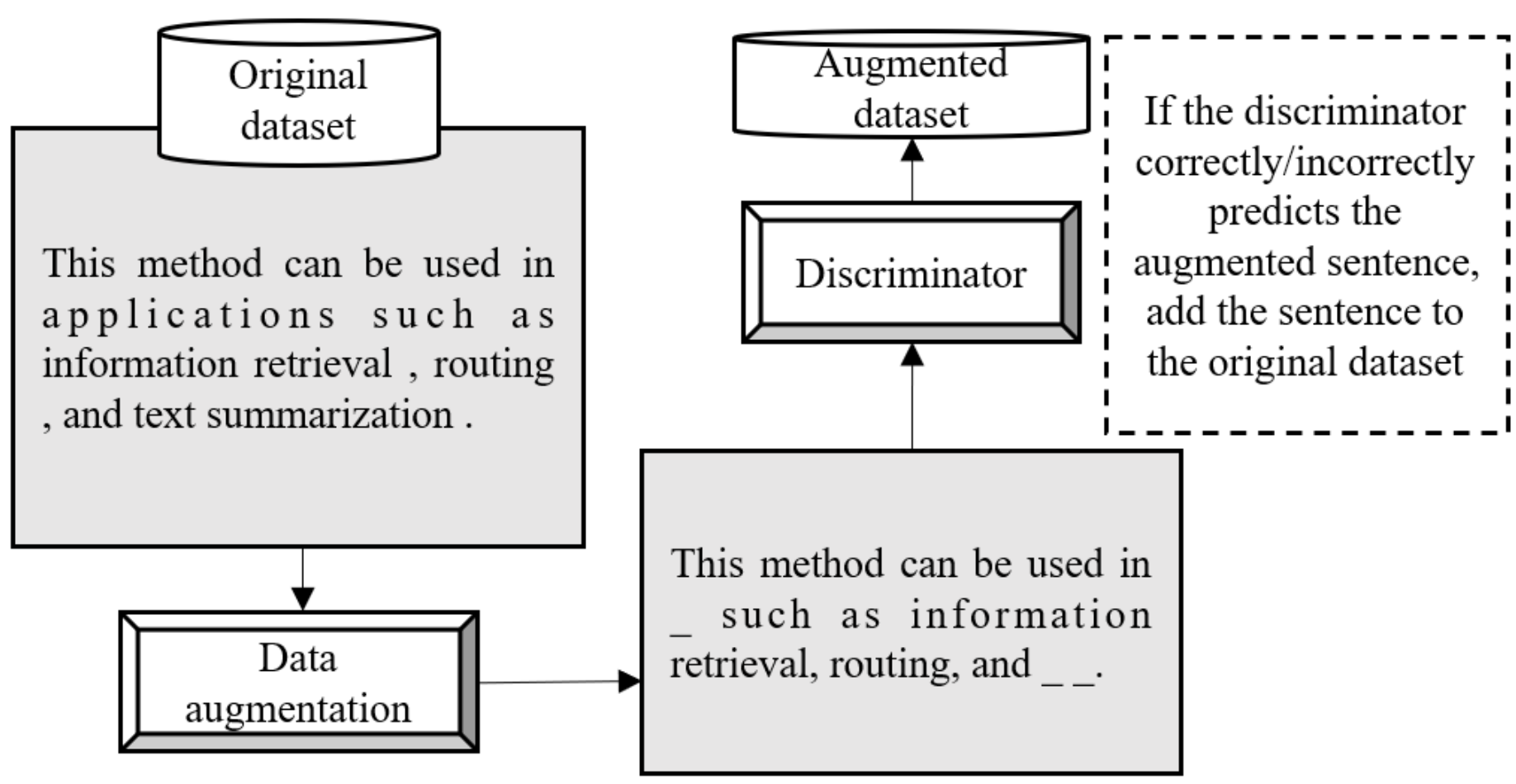

**Figure 2. The framework of sentence quality discrimination methods.**

**3.2 Context-enhanced Transformer**

The context-enhanced transformer consists of three parts, the first is the sentence and context representation module, the second is the context-enhanced transformer module, and the third is the sentence extraction module. We introduce these three modules in detail.

**(1) Sentence and context representation module**

Given a document $P_i$, the sequence of the target sentence $P_{i,s}$ is represented as $\{p_{i,s,1}, p_{i,s,2}, \cdots, p_{i,s,w}, \cdots, p_{i,s,n}\}$. The sequence of the preceding sentence $P_{i,s-1}$ and the following sentence $P_{i,s+1}$ are $\{p_{i,s-1,1}, p_{i,s-1,2}, \cdots, p_{i,s-1,w}, \cdots, p_{i,s-1,n}\}$ and $\{p_{i,s+1,1}, p_{i,s+1,2}, \cdots, p_{i,s+1,w}, \cdots, p_{i,s+1,n}\}$, respectively. We use the symbols "START" and "$END$" to represent the preceding sentence of the first sentence and the following sentence of the final sentence, respectively. The sequence of $P_{i,s}$, $P_{i,s-1}$, and $P_{i,s+1}$ are converted to vector representations $V_{i,s}$, $V_{i,s-1}$, and $V_{i,s+1}$ using a pre-trained language model.

**(2) Context-enhanced transformer module**

To introduce context information, we modify the original structure of the Transformer (Vaswani et al., 2017). In the original transformer, the three inputs are $Query$, $Key$, and $Value$, all of which are the target sentences $V_{i,s}$. Unlike the original transformer, we take the context information as the $Query$ and the target sentence as the $Key$ and $Value$, as shown in the following formulas:

$$Query = W_Q . concat\left(V_{i,s-1}, V_{i,s+1}\right) \tag{1}$$

$$Key = W_K . V_{i,s} \tag{2}$$

$$Value = W_V . V_{i,s} \tag{3}$$

Where the $concat$ represents a concatenation operation, $W_Q$, $W_K$, and $W_V$ represent the weight matrix.

Given a word $v_{i,s',\mathrm{w}}$ in the $Query$, we calculate the dot product of this word with each word in the $Key$. The dot product score represents the importance of each word in the target sentence in terms of the word $v_{i,s',\mathrm{w}}$ in context. Then, the normalized scores are multiplied by $Value$. As shown in the formula:

$$Z = Softmax(QK^T / \sqrt{d_k})V, \tag{4}$$

Where $Q, K, V$ are $Query$, $Key$, and $Value$, respectively. $\sqrt{d_k}$ denotes the scaling factor.

After obtaining $Z$, we get the word representation $S_{i,s,w}$ after operations show in the following formulas.

$$Z' = Norm\ (Z + Value) \tag{5}$$

$$Z'' = Feed - Forward\ (Z') \tag{6}$$

$$S_{i,s,w} = Norm(Z'' + Z') \tag{7}$$

Where $Norm$ and $Feed - Forward$ represent the normalization and the feed-forward layer (Figure 3).

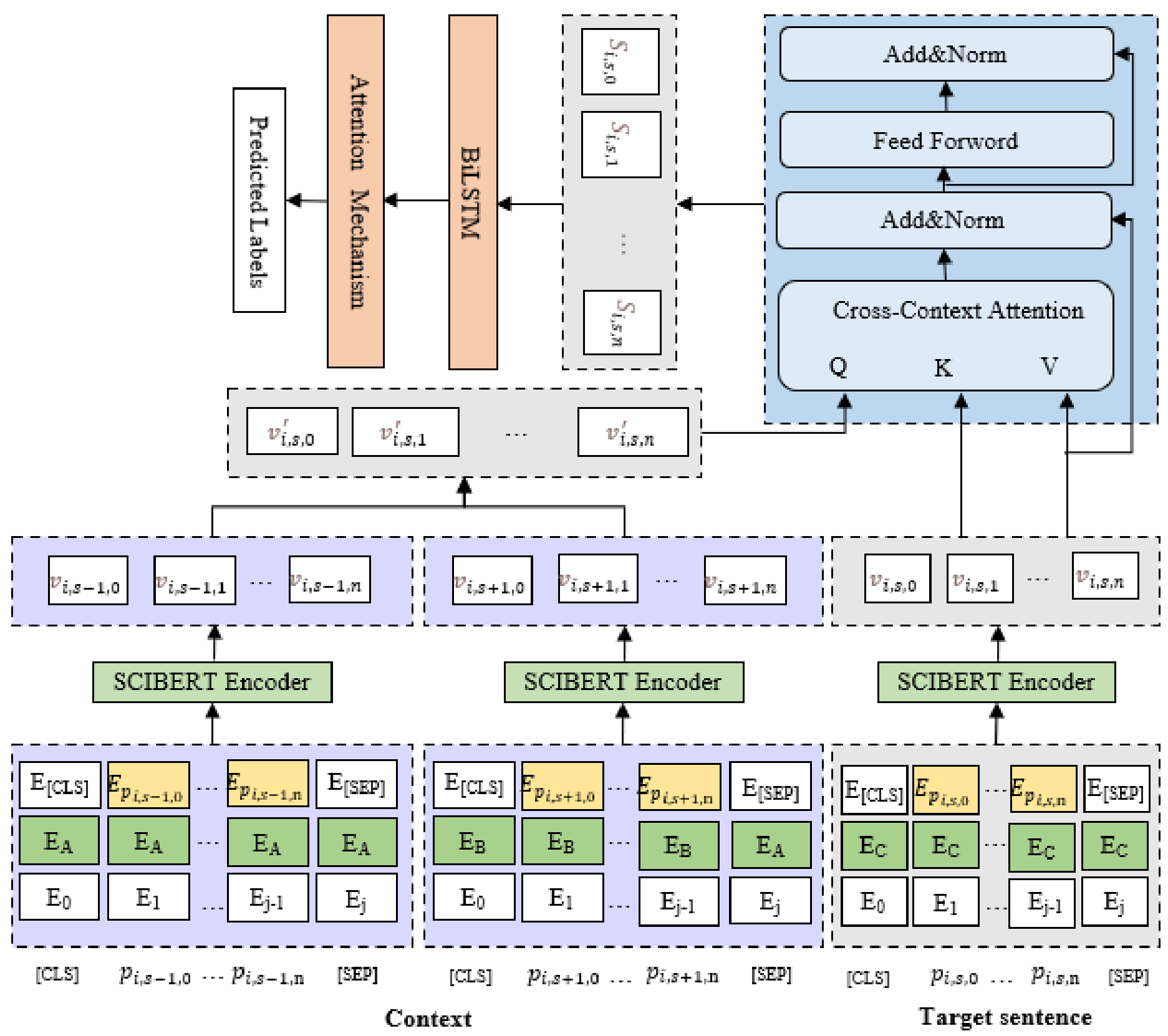

**Figure 3. sentence extraction models with context-enhanced transformer.**

**(3) Sentence extraction module**

This module is used to predict the category of the target sentence with the $S_{i,s}$. In this module, we use BiLSTM (Graves & Schmidhuber, 2005) and attention mechanism (Raffel & Ellis, 2016) to conduct feature learning and then use a softmax layer to conduct prediction. The loss function used is cross entropy.

**3.3 LLM-based In-context learning method**

ICL is a paradigm that allows language models to learn tasks given only a few examples. It is a training-free learning framework that has gained traction with the increasing capabilities of LLMs (Agrawal et al., 2022; Dong et al., 2023). This paper design two LLM-based ICL methods, i.e., sentence-level ICL and paragraph-level ICL. The ICL method concatenates a query question, a piece of demonstration, and other information (e.g., sentence definition) to create a prompt, which is then fed into the LLM for prediction.

- **Sentence-level ICL.** This method directly predicts the category of each sentence. As shown in Figure 4, the prompt consists of the sentence definition, a piece of demonstration, the query question, and the target sentence.

- **Paragraph-level ICL.** This method extracts problem, method, and co-occurrence sentences from a paragraph. As shown in Figure 4, the prompt consists of the sentence definition, a piece of demonstration, the query question, and the target paragraph.

**Prompt of Sentence-level in-context learning**

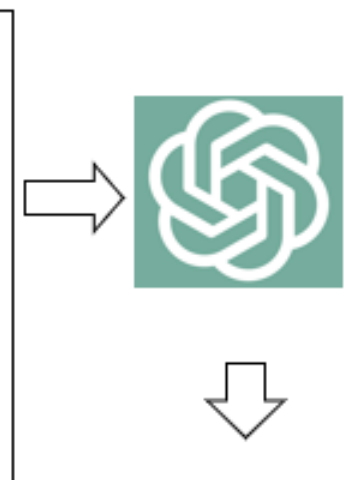

**Prompt of Paragraph-level in-context learning**

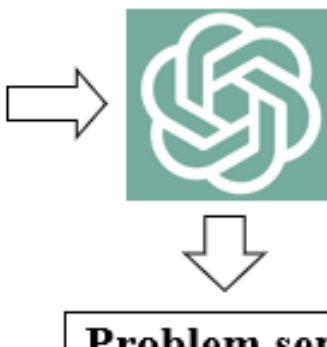

**Figure 4. The framework of sentence-level ICL and paragraph-level ICL.**

# 4. Dataset construction

**Table 3. The definition of three categories of sentences in scientific papers. The underlined values represent the problem or method content in the sentence.**

| Proposed categories | Definition and corresponding example | Source literature | Status of application |
|---|---|---|---|
| Problem sentence | Describe the gap between the methods and the ideal solution<br>**Example:** However this method has a shortcoming: it <u>can not rank the point in the perpendicular of the ideal point and negative ideal point</u>. | Sakai & Hirokawa, 2012 | Exact adoption |
| | Describe the subject of the paper<br>**Example:** This paper considers the problem of <u>automatic assessment of local coherence</u>. | Fisas, Saggion, & Ronzano, 2015; Liakata et al., 2010; Liu et al., 2013; Sakai & Hirokawa, 2012; Yamamoto & Takagi, 2005; | Exact adoption |
| | Describe the subject in past work<br>**Example:** CITE presented a unified framework for <u>natural language disambiguation tasks</u>. | - | New category |
| Method sentence | Describe the process investigators or researchers did that used to achieve a goal<br>**Example:** We introduce a <u>polynomial time</u> | Liakata et al., 2010; Liu et al., 2013; Yamamoto & Takagi, 2005 | Exact adoption |

| | decoding algorithm for the model. | | |
|---|---|---|---|
| | Describe the details of the experimental setup (dataset, metrics, or models)<br>**Example:** We performed our experiments on the last two published RTE datasets: CITE and CITE. | Fisas, Saggion, & Ronzano, 2015; Liakata et al., 2010 | Exact adoption |
| | Describe the method used or mentioned in past work<br>**Example:** CITE used a maximum entropy model considering only lexical information from within the verb phrase (ignoring N2). | Iwatsuki & Aizawa, 2021; Liakata et al., 2010 | Exact adoption |
| cooccurrence sentence | Describe the subject of the paper and the method used to achieve a goal<br>**Example:** We handle the problem of partial observability by inverting the conventional notion of state in a dialogue. | - | New category |
| | Describe the subject and method of past work<br>**Example:** CITE proposed a graph-based semi-supervised model to re-rank n-best translation output. | - | New category |
| | Describe the method and the gap between this method and the ideal solution<br>**Example:** The conceptual retrieval systems, though quite effective, are not yet mature enough to be considered in serious information retrieval applications, the major problems being their extreme inefficiency and the need for manual encoding of domain knowledge (CITE). | - | New category |

**4.1 Sentence definition**

To define method sentences, we adopt definitions from previous studies according to section 2.1. The definitions, along with examples and their corresponding source literature, are given in Table 3.

To define problem sentences, we first adopt two definitions from previous studies. Table 3 represents these definitions and their source literature. Additionally, we add another definition that is "describing the subject in past work". By extracting sentences describing the problem or the method in previous studies, we will have a clear graph depicting the evolution of problems or methods.

Furthermore, we define a new category of sentences: the co-occurrence sentence. This category has been overlooked in previous studies. Conversely, this paper provides three distinct definitions for co-occurrence sentences based on the definition of problem sentences and method sentences. The description and examples of each definition are shown in Table 3.

Finally, any sentences that do not fall into the above three categories are defined as normal sentences. These sentences contain neither problem nor method-related information. It is important to note that some sentences may simply contain general phrases like "method" and "problem." While these phrases are related to problems or methods, they do not represent specific problems or methods. In this paper, such sentences should not be considered as problem or method sentences.

### 4.2 Dataset preparation

This paper constructs two datasets. The first is an Abstract dataset. Abstracts are the core of scientific papers, but they often miss important information. Thus, we also construct a Full-Text dataset.

**Abstract dataset.** This dataset is a subset of the SCIERC dataset (Luan et al., 2018). In SCIERC, there are 500 abstracts selected from twelve artificial intelligence conferences, including natural language processing (NLP). We selected 305 abstracts related to NLP from this dataset. There are 1,440 sentences in this dataset.

**Full-Text dataset.** This dataset consists of 74 papers frpm the Annual Meeting of the Association for Computational Linguistics conference (ACL). The ACL anthology released the XML format of papers published from 1979 to 2015. We randomly sampled two papers each year from the released dataset. There are 12,594 sentences in the dataset.

### 4.3 Cross-weigh-based dataset annotation

Everyone makes mistakes, and so do human annotators (Wang et al., 2019). Previous studies have askd one of the two annotators to annotate the remaining samples after achieving a high Cohen's Kappa score (Pan et al., 2015). Although this approach reduces the labor cost, it will inevitably introduce label errors. To alleviate label errors, we propose a three-step data annotation strategy inspired by Cross-Weigh proposed by Wang et al. (2019).

**Table 4. Statistical information of manual labeling errors and model predicting errors.**

| Dataset | Manual labeling error | | Model predicting error | |
|---|---|---|---|---|
| | quantity | percent | quantity | percent |
| **Abstract** | 111 | 26.0 | 316 | 74.0 |
| **Full-Text** | 589 | 15.22 | 3,280 | 84.78 |

**Step 1: Designing annotation guidelines.** The annotation guidelines are developed based on the sentence definitions provided in section 4.1.

**Step 2: Dataset annotation.** Sentence annotation is conducted in accordance with these guidelines. Following the methodologies employed by Pan et al. (2015) and Zhao, Yan, & Li (2018) , we initially recruited two annotators specializing in NLP, consisting of a master's degree candidate and a doctoral candidate. After training the annotators, we randomly selected 30 samples from the abstract dataset and asked two annotators to annotate them independently. The consistency score measured by Cohen's Kappa is 74.42%, which meets the required level of agreement. Subsequently, the doctoral candidate independently annotate the remaining sentences.

**Step 3: Label error correction.** In order to correct label errors within the annotated datasets, we implemented an error correction framework based on the Cross-Weigh method. First, we applied a ten-fold cross-evaluation method to predict the categories of all sentences within the datasets. For any sentences where the manually annotated labels conflicted with the predicted categories, we categorized these sentences as potential error cases. Then, we engaged a doctoral candidate with expertise in NLP to analyze the potential error set. This set can be further divided into two subsets: one related to manual labeling errors and the other to model prediction errors. The statistics information for these two subsets is shown in Table 4. Finally, our expert addressed the label corrections for the manually labeled errors, resulting in the statistics for the revised dataset provided in Table 5.

**Table 5. Statistical information of two revised datasets.**

| Category \ Dataset | Abstract | Full-Text |
|---|---|---|
| Problem sentence | 271 | 1,160 |
| Method sentence | 395 | 2,428 |
| Cooccurrence sentence | 562 | 2,061 |
| Normal sentence | 212 | 6,945 |

# 5. Experimental results

This section begins with an introduction to the baseline models. It then delves into the parameters and metrics employed for the evaluation. Finally, the results of the proposed models are analyzed.

### 5. 1 Baseline models

First, this paper compares FE desensitization-based data augmenters with several state-of-the-art augmenters:

- **Random augmenter (Random)** randomly selects words and replaces them with the symbol "_" (Wei & Zou, 2019).
- **TFIDF augmenter (TFIDF)** selects words by TFIDF values, randomly choosing words with the top 5 TFIDF values to replace the selected words (Xie et al., 2020).
- **Embedding augmenter (Embed)** uses a random method to select words and then chooses a word from the top 100 most similar words to replace the selected word. GloVe embeddings are used (Wang & Yang, 2015).
- **Contextual embedding augmenter (PLM)** employs a random method to mask words in sentences and then selects a word from the top 100 words with the highest predicted probabilities to replace the masked word (Kobayashi, 2018). The BERT (Devlin et al., 2019) is used.
- **Back translation augmenter** (**BackTrans**) utilizes machine translation to generate synthetic data (Graça et al., 2019). This paper uses the WMT19 En-De model for back translation (Ng et al., 2019).
- **GPT augmenter (GPT)** generates new sentences using LLMs. It first uses a prompt to generate method and problem entities and then generates sentences with these generated entities (Ding et al., 2022).

Second, this paper compares the context-enhanced transformer with state-of-the-art sentence extraction models, some of which use context information, while others do not:

- **PLM-BiLSTM-ATT** consists of a pre-trained language model, a BiLSTM, and an attention mechanism. The context is not considered in this model (Zhang & Ren, 2020).
- **PLM-BiLSTM-TRANS** includes a pretrained language model, a BiLSTM, and a transformer (Chen et al., 2022; Zhao et al., 2020). Similar to PLM-BiLSTM-ATT, context is not used in the model.
- **Sequential model** involves a pre-trained language model and a hierarchical BiLSTM with an attention mechanism (Jin & Szolovits, 2018). It predicts the labels of all sentences in a sequence simultaneously. Given the large number of sentences in a scientific paper, this paper used the target sentence, preceding sentence, and following sentence to construct a sequence.

- **Concatenation model** uses a pretrained language model to encode both the context and the target sentence (Zhang et al., 2020). The encoded values are then directly concatenated. This concatenated representation is fed into a BiLSTM and an attention layer.

## 5. 2 Parameter settings

The paper provides specific parameter settings for data augmentation, sentence extraction models, and LLM-based ICL:

**Data Augmentation Parameter Settings:** The number of words replaced in the FE desensitization-based data augmenters and their baselines should not exceed 20% of the total words in the original sentence. The PLM-BiLSTM-ATT model is employed to train the quality discriminator for data augmentation.

**Sentence Extraction Model Parameter Settings:** For the context-enhanced transformer and its baselines, the Adam optimizer is applied. The number of epochs is set to 20 for the abstract dataset and 10 for the full-text dataset. The learning rate is set to 3e-5. The maximum sentence length is limited to 512 tokens. The number of neurons in the BiLSTM is set to 150. In the context-enhanced transformer, an eight-head self-attention mechanism is used.

**LLM-based ICL Parameter Settings:** The GPT-3.5 model is chosen as the backbone for this method. The "gpt-3.5-turbo" API provided by OpenAI is used to access GPT-3.5.

## 5. 3 Evaluation metrics

The accuracy, precision ($P$), recall ($R$), $F_1$, macro average precision ($Macro\ P$), macro average recall ($Macro\ R$), and macro average $F_1$ ($Macro\ F_1$) are used to evaluate the performance of models. To denote the $F_1$ scores for specific types of sentence extraction, we use the following abbreviations: $F_1$@T for problem sentence extraction, $F_1$@M for method sentence extraction, $F_1$@MT for cooccurrence sentence extraction, and $F_1$@N for normal sentence extraction.

## 5. 4 Performance of proposed models

### 5.4.1 Pretrained language model selection

To select an appropriate pre-trained language model, we conducted a performance comparison among three models, i.e., BERT, RoBERTa (Liu et al., 2019), and SCIBERT.

Table 6 presents the performance results of PLM-BiLSTM-ATT with different pretrained language models on two datasets. It is evident that models utilizing SCIBERT achieve superior macro $F_1$ scores, reaching 70.62% in the abstract dataset and 71.46% in the full-text dataset, outperforming those using BERT and RoBERTa. Consequently, we choose for SCIBERT as the pretrained language model for subsequent experiments.

**Table 6. Results of three pretrained language models on two datasets. The values in the Table are expressed as a percentage (%). The bold font values indicate the best performance on each metric.**

| **Model** \ **Metric** | | Acc | Macro P | Macro R | Macro $F_1$ | $F_1$@ T | $F_1$@ M | $F_1$@ MT | $F_1$@ N |
|---|---|---|---|---|---|---|---|---|---|
| **Abstract** | BERT | 69.99 | 70.98 | 67.34 | 67.93 | **65.19** | 66.65 | 74.67 | 65.22 |

| | RoBERTa | 71.13 | 71.14 | 69.38 | 69.56 | 62.68 | 69.71 | 75.27 | **70.56** |
|---|---|---|---|---|---|---|---|---|---|
| | SCIBERT | **73.18** | **72.42** | **70.50** | **70.62** | 63.46 | **73.50** | **77.57** | 67.96 |
| **Full-Text** | BERT | 78.12 | 72.14 | 70.84 | 70.99 | **60.37** | 64.76 | **71.60** | 87.23 |
| | RoBERTa | 76.99 | 69.66 | 69.04 | 69.09 | 55.81 | 63.99 | 69.59 | 86.96 |
| | SCIBERT | **78.77** | **72.39** | **71.05** | **71.46** | 60.14 | **66.17** | 71.56 | **87.95** |

#### 5.4.2 Performance of FE-desensitization based data augmenters

In FE desensitization-based data augmenters, we employ three strategies to select FEs. Table 7 illustrates 10 FEs that have been selected using these three strategies.

**Table 7. 10 FEs selected by three FE selection strategies**

| FE selection strategy | FEs |
|---|---|
| Dependency-tree based strategy | of, for, to, system, model, use, in, with, through, from |
| Open-source based strategy | is the lack of， we present a， this paper presents， we describe two， in this paper we adopt a, in this paper we aim at, we study the problem of, in this paper we study whether, this allows the, are based on a |
| LLM based strategy | can be applied to, to show the effectiveness of the, have focused on, according to, presents a, is proposed, For this purpose, In order to, based on, is implemented as |

Subsequently, we evaluate the performance of FE desensitization-based data augmenters on two datasets. As illustrated in Table 8, we draw the following conclusions:

**(1) In the abstract and full-text dataset, the FE-desensitization-based data augmenter with a dependency-tree-based FE selection and LLM-based FE selection strategy yields the best macro $F_1$, respectively.** Specifically, the best macro $F_1$ in the abstract dataset is 73.98%, and in the full-text dataset, it is 73.36%. These scores surpass those achieved by the PLM-BiLSTM-ATT model by 3.36% and 1.9%, respectively. The dependency-tree-based and LLM-based FE selection strategy extract FEs from the target training dataset, whereas the open-source-based FE selection strategy uses FEs provided by Iwatsuki and Aizawa (2021). This observation indicates that utilizing FEs extracted from target datasets enhances the performance of the FE-desensitization-based data augmenter.

**(2) In the abstract and full-text dataset, the augmenter with the "true-discrimination" and "wrong-discrimination" strategies achieves the best macro $F_1$, respectively.** For instance, in the abstract dataset, the macro F1 of the random augmenter with the "true-discrimination" strategy is 1.62% higher than that of the random augmenter with the "wrong-discrimination" strategy. In the full-text dataset, the macro F1 of the random augmenter with the "true-discrimination" strategy is 6.66% lower than that of the random augmenter with the "wrong-discrimination" strategy. We will provide a detailed analysis of this phenomenon in section 6.1.

**Table 8. Results of data augmenters. "+Random" denotes the random augmenter combined with PLM-BiLSTM-ATT. "DependTree FE", "LLM FE", and "OpenSource FE" denote the dependency-tree-based, LLM-based, and open-source-based FE selection strategies. "True" indicates the "true-discrimination" strategy , while "False" represents the "wrong-discrimination" strategy. The remaining details can be inferred.**

| **Model** \ **Metric** | | Acc | Macro P | Macro R | Macro $F_1$ | $F_1$@ T | $F_1$@ M | $F_1$@ MT | $F_1$@ N |
|---|---|---|---|---|---|---|---|---|---|
| | PLM-BiLSTM-ATT | 73.18 | 72.42 | 70.50 | 70.62 | 63.46 | 73.50 | 77.57 | 67.96 |

| | | | | | | | | |
|---|---|---|---|---|---|---|---|---|
| | +Random-True | 73.80 | 74.17 | 72.53 | 72.25 | 67.66 | 72.65 | 77.77 | 70.92 |
| | +Random-False | 72.74 | 74.07 | 70.59 | 70.63 | 64.37 | 72.17 | 76.20 | 69.77 |
| | +TFIDF-True | 73.72 | 73.40 | 71.05 | 71.26 | 65.35 | 73.96 | 78.17 | 67.55 |
| | +TFIDF- False | 71.52 | 73.11 | 68.42 | 68.95 | 63.67 | 71.28 | 76.35 | 64.49 |
| | +Embed-True | 73.47 | 73.50 | 72.73 | 72.07 | 66.03 | 74.22 | 75.75 | 72.27 |
| | +Embed- False | 71.78 | 72.94 | 69.24 | 69.76 | 63.18 | 71.72 | 75.86 | 68.29 |
| | +PLM-True | 74.52 | 75.37 | 72.21 | 72.84 | 66.01 | 74.12 | 78.29 | 72.95 |
| | +PLM-False | 70.42 | 70.90 | 69.81 | 69.22 | 61.72 | 71.22 | 73.31 | 70.61 |
| | +BackTrans-True | 75.18 | 75.60 | 72.08 | 72.88 | 67.32 | 74.28 | 78.95 | 70.95 |
| | +BackTrans-False | 72.71 | 73.80 | 70.48 | 70.87 | 64.45 | 72.64 | 76.69 | 69.69 |
| | +ChatGPT Gene-True | 74.03 | 74.91 | 71.84 | 71.96 | **68.36** | 72.92 | 78.65 | 67.93 |
| | +ChatGPT Gene- False | 69.74 | 70.00 | 68.07 | 67.80 | 64.73 | 68.03 | 73.85 | 64.61 |
| | +DependTree FE-True | 75.73 | 76.88 | **73.54** | **73.98** | 67.22 | 75.67 | 79.32 | 73.70 |
| | +DependTree FE -False | 77.68 | 71.33 | 69.66 | 70.08 | 57.96 | 64.33 | 70.70 | 87.35 |
| | +LLM FE-True | 73.57 | 73.56 | 71.69 | 72.06 | 66.21 | 72.03 | 76.67 | 73.34 |
| | +LLM FE-False | 71.03 | 70.94 | 69.93 | 69.73 | 63.29 | 70.56 | 74.84 | 70.22 |
| | +OpenSource FE-True | 76.07 | **76.89** | 72.57 | 73.64 | 67.14 | **77.22** | **79.77** | 70.43 |
| | +OpenSource FE-False | **79.14** | 73.05 | 72.10 | 72.34 | 61.27 | 67.21 | 72.75 | **87.94** |
| **Full Text** | PLM-BiLSTM-ATT | 78.77 | 72.39 | 71.05 | 71.46 | 60.14 | 66.17 | 71.56 | 87.95 |
| | +Random-True | 76.11 | 66.31 | 65.57 | 65.48 | 47.15 | 60.03 | 67.99 | 86.73 |
| | +Random-False | 78.61 | 72.16 | 72.76 | 72.14 | 61.36 | 66.84 | 73.06 | 87.30 |
| | +TFIDF-True | 78.81 | 72.93 | 70.12 | 71.12 | 58.65 | 66.31 | 71.55 | 87.96 |
| | +TFIDF- False | 78.26 | 71.83 | 71.27 | 71.24 | 60.38 | 65.73 | 71.43 | 87.40 |
| | +Embed-True | 76.48 | 68.39 | 67.70 | 66.98 | 53.11 | 58.33 | 69.44 | 87.04 |
| | +Embed- False | 79.34 | 72.92 | 72.84 | 72.71 | 62.09 | 66.81 | 74.15 | 87.79 |
| | +PLM-True | 77.82 | 71.61 | 69.41 | 70.03 | 56.94 | 65.07 | 70.62 | 87.49 |
| | +PLM-False | 77.84 | 71.64 | 70.58 | 70.56 | 59.41 | 65.04 | 70.60 | 87.17 |
| | +BackTrans-True | 78.84 | 72.13 | 72.44 | 71.96 | 60.15 | 67.31 | 72.44 | 87.95 |
| | +BackTrans-False | 78.75 | 72.05 | 72.61 | 71.99 | 60.70 | 67.86 | 71.77 | 87.62 |
| | +ChatGPT Gene-True | 79.59 | 73.74 | 72.13 | 72.80 | 62.14 | 67.52 | 72.20 | **89.32** |
| | +ChatGPT Gene-False | 79.51 | 73.00 | **73.46** | 72.77 | 61.63 | **67.92** | 73.25 | 88.26 |
| | +DependTree FE-True | 77.68 | 71.33 | 69.66 | 70.08 | 57.96 | 64.33 | 70.70 | 87.35 |
| | +DependTree FE -False | 79.67 | 73.41 | 73.24 | 73.11 | 62.20 | 67.72 | **74.54** | 87.99 |
| | +LLM FE-True | 78.89 | 72.26 | 71.41 | 71.56 | 59.27 | 66.75 | 72.19 | 88.02 |
| | +LLM FE-False | **79.77** | **74.03** | 73.27 | **73.36** | **63.60** | 67.56 | 74.12 | 88.16 |
| | +OpenSource FE-True | 79.14 | 73.05 | 72.10 | 72.29 | 61.27 | 67.21 | 72.75 | 87.94 |
| | +OpenSource FE-False | 79.53 | 73.55 | 72.76 | 72.83 | 62.29 | 67.69 | 73.10 | 88.22 |

### 5.4.3 Performance of context-enhanced transformer

This section analyzes the results of the context-enhanced transformer and its baselines on two datasets. According to Table 9, we can draw the following conclusions:

**(1) The macro $F_1$ of the context-enhanced transformer is higher than that of PLM-BiLSTM-ATT.** In both the abstract and full-text dataset, the macro $F_1$ of the context-enhanced transformer is 73.45% and 72.98%, respectively, which is 2.83% and 1.52% higher than PLM-BiLSTM-ATT.

**(2) The macro $F_1$ of the context-enhanced transformer outperforms the concatenation model.** In both the abstract and full-text dataset, the macro $F_1$ of the context-enhanced transformer is higher than the concatenation model by 1.64% and 0.44%, respectively. The concatenation model, which directly combines context and target sentence representations, can introduce noise into the model since not all context content is necessarily relevant to the target sentence. In contrast, the context-enhanced transformer effectively filters out noise from the context while enhancing the relevant information.

**(3) Sequential models are not suitable for the problem and method sentence extraction task.** The sequential model yields a lower macro $F_1$ compared to other models in both datasets. This is due to the sparsity of problem and method sentences in scientific papers, making the sequential model less effective for tasks with sparse labels.

**Table 9. Results of the context-enhanced transformer and its baselines.**

| | Model \ Metric | Acc | Macro P | Macro R | Macro $F_1$ | $F_1$@T | $F_1$@M | $F_1$@MT | $F_1$@N |
|---|---|---|---|---|---|---|---|---|---|
| Abstract | PLM-BiLSTM-ATT | 73.18 | 72.42 | 70.50 | 70.62 | 63.46 | **73.50** | 77.57 | 67.96 |
| | PLM-BiLSTM-TRANS | 73.21 | 72.64 | 72.21 | 71.67 | 65.09 | 72.52 | 76.70 | 72.37 |
| | Sequential model | 43.09 | 52.54 | 40.10 | 35.29 | 15.73 | 35.00 | 55.36 | 35.08 |
| | Concatenation model | 73.23 | 73.22 | 72.44 | 71.81 | 65.63 | 73.07 | 75.99 | 72.53 |
| | Context-enhanced Transformer | **74.85** | **74.75** | **74.02** | **73.45** | **68.68** | 73.37 | **77.95** | **73.79** |
| Full-Text | PLM-BiLSTM-ATT | 78.77 | 72.39 | 71.05 | 71.46 | 60.14 | 66.17 | 71.56 | 87.95 |
| | PLM-BiLSTM-TRANS | 78.98 | 72.61 | 72.53 | 72.21 | 60.62 | 67.11 | 73.30 | 87.79 |
| | Sequential model | 68.21 | 61.38 | 49.84 | 49.81 | 15.16 | 47.80 | 54.67 | 81.59 |
| | Concatenation model | 79.16 | 72.46 | 73.16 | 72.54 | 61.67 | **67.40** | 73.13 | 87.97 |
| | Context-enhanced Transformer | **79.87** | **73.50** | **73.05** | **72.98** | **62.06** | 67.24 | **74.12** | **88.48** |

**5.4.4 Performance of context-enhanced transformer combined with augmenters**

To answer RQ3, in this section, we combined the context-enhanced transformer with data augmenters using both the “true-discrimination” and “wrong-discrimination” strategies in the abstact and full-text datsests, respectively, Building on the results from section 5.4.2, Table 10 reveals the following insights:

**(1) In both datasets, the context-enhanced transformer combined with the FE augmenter using dependency-tree-based FE selection strategy achieves the best macro $F_1$.** In the abstract dataset, this combined model achieves a macro $F_1$ of 74.33%, which is 3.71% higher than that of PLM-BiLSTM-ATT and 0.88% higher than the context-enhanced transformer. In the full-text dataset, this combined model achieves a macro $F_1$ of 74.13%, which is 2.67% higher than PLM-BiLSTM-ATT and 1.15% higher than the context-enhanced transformer.

**(2) Combining the context-enhanced transformer with augmenters yields a better macro $F_1$ than individual models.** For example, in the abstract dataset, the macro $F_1$ of the context-enhanced transformer combined with the random augmenter is 73.94%, which is 0.49% higher than the context-enhanced transformer and 1.69% higher than the random augmenter.

**Table 10. The results of the combined models. “*” indicates that the macro $F_1$ of the context-enhanced Transformer combined with the augmenter and the PLM-BiLSTM-ATT are significantly different.**

| | Model \ Metric | Acc | Macro P | Macro R | Macro $F_1$ | $F_1$@T | $F_1$@M | $F_1$@MT | $F_1$@N |
|---|---|---|---|---|---|---|---|---|---|
| Abstract | Context-enhanced Transformer | 74.85 | 74.75 | 74.02 | 73.45 | 68.68 | 73.37 | 77.95 | 73.79 |
| | +Random-True | 75.18 | **76.21** | 73.33 | 73.94 | **70.18** | 73.29 | 78.28 | 74.04 |
| | +TFIDF-True | 75.17 | 75.14 | 73.86 | 73.92 | 68.59 | 74.26 | 78.97 | 73.85 |
| | +Embed-True | **78.77** | 75.50 | 73.35 | 73.91 | 66.28 | 68.75 | 75.74 | **84.86** |
| | +PLM-True | 75.42 | 75.44 | 73.05 | 73.57 | 67.12 | 73.98 | **79.90** | 73.26 |
| | +ChatGPT-True | 75.02 | 74.67 | 74.04 | 73.58 | 70.02 | 73.16 | 78.61 | 72.52 |
| | +BackTrans-True | 75.27 | 75.65 | 73.38 | 73.59 | 68.45 | 75.04 | 79.05 | 71.82 |
| | +DependTree FE-True | 76.09 | 75.65 | 74.14 | **74.33**$^{*}$ | 69.00 | 74.92 | 79.72 | 73.67 |
| | +ChatGPT FE-True | 75.25 | 75.30 | **74.47** | 74.14* | 68.63 | 73.43 | 78.47 | 76.05 |
| | +OpenSource FE-True | 75.72 | 75.63 | 74.12 | 74.12* | 66.55 | **76.46** | 79.44 | 74.04 |
| **Full Text** | Context-enhanced Transformer | 79.87 | 73.50 | 73.05 | 72.98 | 62.06 | 67.24 | 74.12 | **88.48** |
| | +Random-False | 79.65 | 73.89 | 72.84 | 73.08 | 62.49 | 68.10 | 73.65 | 88.07 |
| | +TFIDF-False | 79.71 | 73.25 | 73.21 | 73.05 | 62.56 | 68.15 | 73.19 | 88.31 |
| | +Embed-False | 79.73 | 74.00 | 72.95 | 73.17 | 62.22 | 68.08 | 74.23 | 88.15 |
| | +PLM-False | 79.89 | 74.01 | 73.05 | 73.34 | 63.04 | 68.13 | 74.10 | 88.09 |
| | +ChatGPT-True | 79.68 | 73.83 | 72.69 | 73.03 | 62.65 | 67.89 | 73.29 | 88.27 |
| | +BackTrans-False | 79.52 | 73.55 | 73.84 | 73.37 | 64.02 | 68.11 | 73.49 | 87.86 |
| | +DependTree FE-True | **80.38** | **74.50** | **74.14** | **74.13*** | **64.10** | 68.76 | **75.31** | 88.36 |
| | +ChatGPT FE-True | 79.49 | 73.42 | 73.35 | 73.23* | 63.92 | 67.75 | 73.35 | 87.89 |
| | +OpenSource FE-True | 79.87 | 73.81 | 73.69 | 73.59* | 63.13 | **69.02** | 74.18 | 88.05 |

#### 5.4.5 Performance of LLM based in context learning

To answer RQ4, this paper compares the macro $F_1$ of LLM-based ICL and PLM-BiLSTM-ATT. As demonstrated in Table 11, both LLM-based ICL methods are found to be unsuitable for the tasks of problem and method sentence extraction. For instance, in both the abstract dataset and the full-text dataset, the macro $F_1$ of the sentence-level ICL are significantly lower than those of PLM-BiLSTM-ATT. Further error analysis of the results of LLM-based ICL methods will be conducted in section 6.3.

**Table 11. Results of two LLM based ICL methods. "*" indicates that the macro $F_1$ of the LLM-based ICL method and the PLM-BiLSTM-ATT are significantly different.**

| | Model \ Metric | Acc | Macro P | Macro R | Macro $F_1$ | $F_1$@ T | $F_1$@ M | $F_1$@ MT | $F_1$@ N |
|---|---|---|---|---|---|---|---|---|---|
| **Abstract** | PLM-BiLSTM-ATT | **73.18** | **72.42** | **70.50** | **70.62*** | **63.46** | **73.50** | **77.57** | **67.96** |
| | Sentence-level ICL | 32.02 | 24.14 | 26.20 | 21.97 | 3.18 | 36.21 | 38.65 | 9.83 |
| | Paragraph-level ICL | 25.85 | 25.96 | 26.57 | 23.07 | 13.13 | 37.88 | 21.20 | 20.08 |
| **Full-Text** | PLM-BiLSTM-ATT | **78.77** | **72.39** | **71.05** | **71.46*** | **60.14** | **66.17** | **71.56** | **87.95** |
| | Sentence-level ICL | 37.80 | 23.17 | 26.65 | 22.77 | 2.24 | 16.05 | 23.36 | 49.41 |
| | Paragraph-level ICL | 32.39 | 26.99 | 27.25 | 21.70 | 16.77 | 13.32 | 5.08 | 51.65 |

## 6. Discussion

This section delves into the mechanisms underlying the functioning of the FE-desensitization based data augmenters and the context-enhanced transformer. Additionally, it includes an error analysis of the LLM-based ICL methods.

#### 6.1 The mechanism of FE desensitization based data augmenter

In order to answer RQ1, this section conducts a comparison of word importance in sentences between the FE desensitization-based data augmenters and the baseline augmenters. The importance of words is represented using the outputs of the attention mechanism in the models. Furthermore, to gain insights into the varying effects of the data discrimination strategy on the abstract dataset and the full-text dataset, as presented in Section 5.4.2, this paper generates training datasets with differing sample sizes and analyzes the influence of the data discrimination strategies on these datasets.

**(1) The comparision of the importance degree of each word in sentence**

To analyze the patterns that these data augmenters have reduced the model's dependence on, this paper calculates the intersection between the high-attention word set of PLM-BiLSTM-ATT and the low-attention word set of data augmenters. For comparison, the FE desensitization-based data augmenters are compared with baseline augmenters, including the random augmenter and the back translation augmenter. The high-attention word set comprises the four words with the highest attention values, while the low-attention word set comprises the four words with the lowest attention values.

To check the robustness of the FE desensitization-based data augmenter, we randomly selects 50, 100, 200, and 500 samples from each fold of the abstract and full-text training datasets. Tables 12 and 13 display the intersection results for each dataset. The following conclusions can be drawn: **Compared with baseline augmenters, the FE desensitization-based data augmenter can reduce the models' dependence on more FE-related words.** These words can be categorized into two categories: one includes words related to problems and methods, such as "task," "framework," and "problem," and the other includes words that connect problem and method content, such as "to," "on," and "with." Since sentences containing these words may not necessarily be problem or method sentences, reducing the importance of these words will improve precision in problem and method sentence extraction.

**Table 12. The intersection results between high-attention word set of PLM-LSTM-ATT and low-attention word set of data augmenter in abstract dataset. The bold words indicate the unique words in all results.**

| Augmenter \ Sample | 50 | 100 | 200 | 500 |
|---|---|---|---|---|
| +Random | of | a; is; paper | of; model; ambiguity | of; for; system |
| +BackTrans | of | A; is; paper | model; ambiguity | paper; which; different |
| +DependTree FE | is; paper; **have;** system | **to; in;** the; **patterns** | a; is; of; **are;** model | **to; in;** of; **on;** for |
| +LLM FE | **and; the;** for | of; **to; in; the** | a; is; of | by; **to;** of; model; **problem** |
| +OpenSource FE | of; **both** | **as;** of | **language; task** | a; the; for; **elements; language** |

**Table 13. The intersection results between high-attention word set of PLM-LSTM-ATT and low-attention word set of data augmenter in full-text dataset. The bold words indicate the unique words in all results.**

| Augmenter \ Sample | 50 | 100 | 200 | 500 |
|---|---|---|---|---|
| +Random | a; as; or; and; the | a; we; in; the; this | a; of; for; the; system | is; the; user; from; system |
| +BackTrans | a; to; on; the; for | a; in; of; to; the | the; example | of; is; to; the; from |
| +DependTree FE | a; of; and; are; the | of; in; to; for; **information** | of; the; and; **framework** | of; and; the; **user; dialogue** |
| +LLM FE | **are;** the; for; this | a; to; of; for; **with** | is; of; the; **model;** system | of; to; user; **them; approach** |
| +OpenSource FE | it; to; and; the; from | of; to; in; for; the | of; in; and; the; **our** | of; to; in; system; **problem** |

To delve deeper into the mechanism by which FE desensitization-based data augmenters operate, we present a case study. We revisit the instance in Table 1, with detailed information provided in Table 14. Among them, the predictions of the PLM-BiLSTM-ATT and random augmenters are co-occurrence sentences. In contrast, the predicted result of FE desensitization-based data augmenters is method sentences, which is same as the result of human annotation.

**Table 14. Case of the FE desensitization-based data augmenter. The bold word represents the FE in this sentence. "CITE" denotes the citation information.**

| sentence | Prediction results |
|---|---|
| More specifically, we combine a probabilistic topological field parser **for** German CITE with the HPSG parser of CITE. | **Human annotation:** method sentence<br>**PLM-BiLSTM-ATT**: cooccurrence sentence<br>**+Random**: cooccurrence sentence<br>**+DependTree FE:** method sentence<br>**+LLM FE:** method sentence<br>**+OpenSource FE:** method sentence |

To further investigate the ability of FE desensitization-based data augmenters, we compared the output of the attention mechanism across models using various data augmenters. As depicted in Figure 5, the attention values for the word "for" generated by PLM-BILSTM-ATT with the dependency-tree-based strategy, LLM-based strategy, and open-source-based strategy are 0.44, 0.45, and 0.67, respectively. These values are all lower than those of PLM-BILSTM-ATT and the random augmenter. This observation provides empirical evidence that FE desensitization-based data augmenters effectively diminish the significance of FEs within sentences.

| Model | more | specifically | , | we | combine | a | probabilistic | topological | field | parser | for | german |
|---|---|---|---|---|---|---|---|---|---|---|---|---|
| PLM-BiLSTM-ATT | 0 | 8.5e-05 | 0.0055 | 0.041 | 0.12 | 0.22 | 0.29 | 0.37 | 0.46 | 0.55 | 0.71 | 0.59 |
| Random | 0 | 7e-05 | 0.0026 | 0.017 | 0.11 | 0.28 | 0.41 | 0.6 | 0.91 | 1 | 0.8 | 0.6 |
| DependTree FE | 0 | 1.2e-05 | 4.9e-06 | 0.011 | 0.059 | 0.12 | 0.18 | 0.23 | 0.3 | 0.34 | 0.44 | 0.49 |
| LLM FE | 0 | 1.7e-05 | 0.0018 | 0.017 | 0.059 | 0.12 | 0.18 | 0.24 | 0.32 | 0.38 | 0.45 | 0.5 |
| OpenSource FE | 0 | 8.2e-06 | 0.0019 | 0.029 | 0.13 | 0.25 | 0.38 | 0.48 | 0.58 | 0.64 | 0.67 | 0.85 |

Sentence

**Figure 5. Heatmap of the attention mechanism outputs in five models. Due to space limitations, only the first half of the sentence is shown.**

**(2) Analysis of the impact of data discrimination strategy**

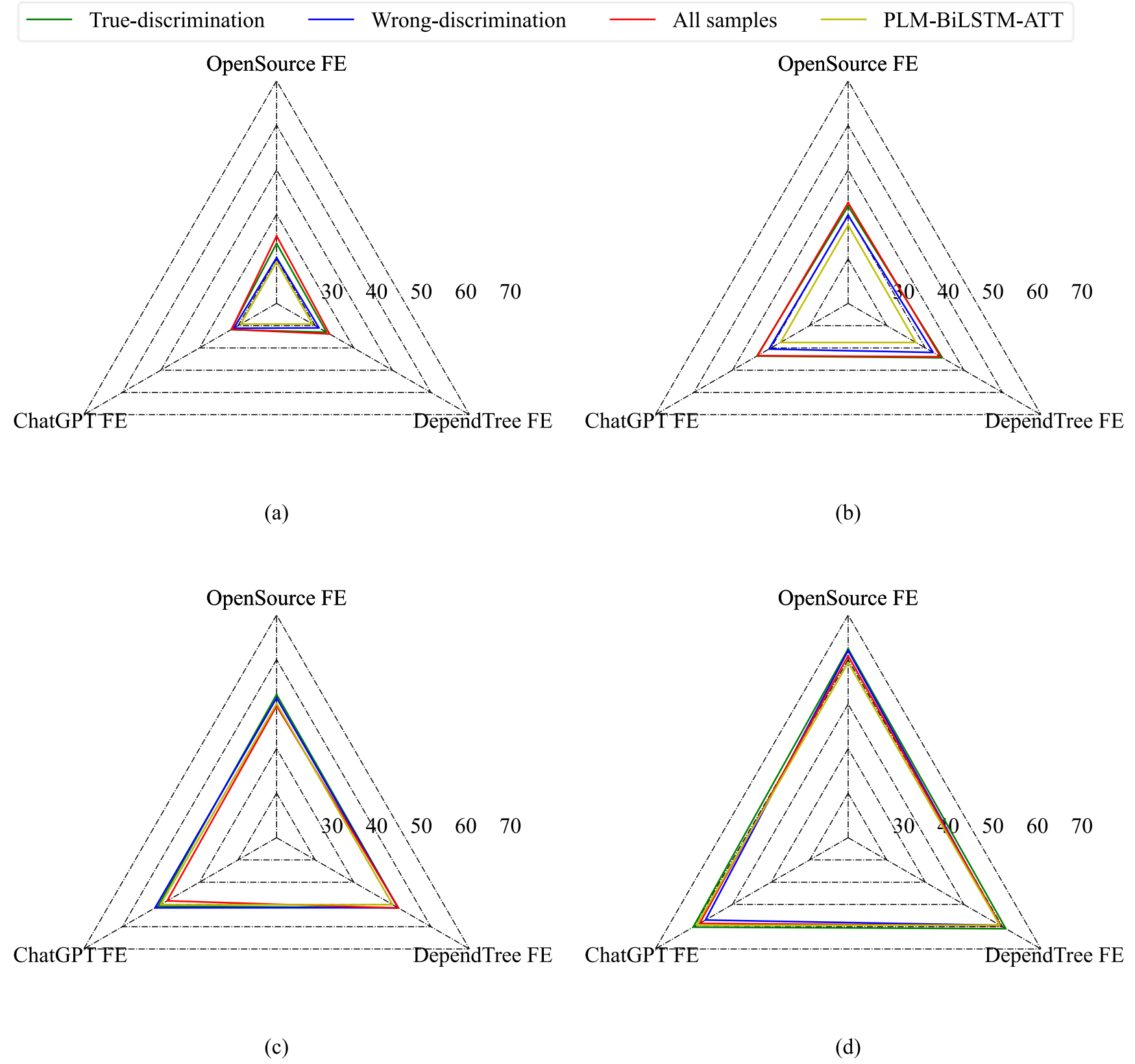

**Figure 6. (a), (b), (c), and (d) represents the macro $F_1$ of FE desensitization based data augmenters on abstarct dataset with sample sizes of 50, 100, 200, and 500, respectively. The green line and blue line represent the data augmenters using the "true-discrimination" and "wrong-discrimination" strategy, respectively. The red line represents the data augmenters that does not employ a data discrimination strategy. The yellow line represents the macro $F_1$ of the PLM-BiLSTM-ATT without augmenters. The same applies to the figures below.**

This section analyzes the impact of "true-discrimination" and "wrong-discrimination" strategies on the models. We randomly selects 50, 100, 200, and 500 samples from each fold of the abstract and full-text training datasets. Using these samples, we trained sentence extraction models with three FE desensitization-based data augmenters and two types of data discrimination strategies. Additionally, we included all augmented sentences in the training dataset to further analyze the necessity of the data discrimination strategy. Figures 6 and 7 display the macro $F_1$ of the models. From these figures, we can observe the following:

**In the abstract datasets, when the dataset is small, the macro $F_1$ of using the "true-discrimination" strategy and including all augmented sentences is comparable. As the dataset size grows, the macro $F_1$ of using the "true-discrimination" strategy outperforms other strategies.** This result aligns with the findings in Section 5.4. To explain this phenomenon, we analyzed the label distribution of samples selected by the "true-discrimination" strategy and the "wrong-discrimination" strategy, respectively. It was noted that the samples selected by the "wrong-discrimination" strategy contained a higher proportion of problem sentences. This observation can clarify the phenomenon: in the abstract dataset, the number of sentences in each category is balanced. Adding a significant number of problem sentences would disrupt the distribution of sentence categories.

**In the full-text datasets, when the dataset is small, the macro $F_1$ of using the "true-discrimination" strategy is higher than other strategies. As the dataset size gradually increases, the macro $F_1$ of using the "wrong-discrimination" strategy becomes the best.** This result is consistent with the findings in Section 5.4.2. The observation of the label distribution of samples selected by the "wrong-discrimination" strategy offers an explanation for this phenomenon. In the full-text dataset, there are significant differences in the number of sentences in different categories, with normal sentences being the most abundant and problem sentences the least. When the dataset size is 50, adding sentences with incorrect predictions introduces noise. However, as the dataset size increases, adding problem sentences to the training dataset enhances the model's ability to recognize problem sentences, thereby improving the model's performance.

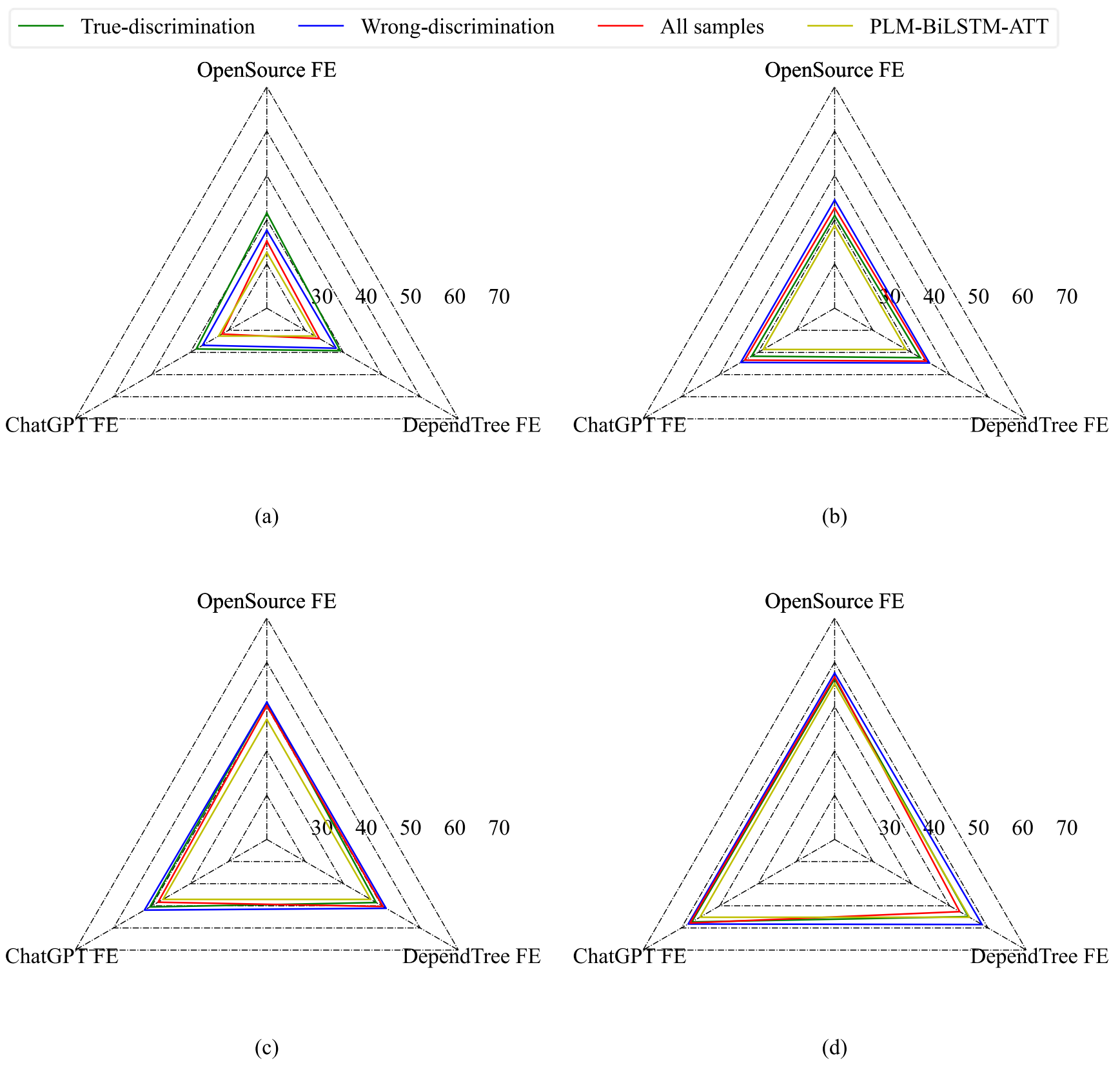

**Figure 7. (a), (b), (c), and (d) represents the macro $F_1$ of FE desensitization based data augmenters on full-text dataset with sample sizes of 50, 100, 200, and 500, respectively.**

### 6.2 The mechanism of context-enhanced transformer

To answer RQ2, we examine the influence of context on the importance of words within sentences. Subsequently, to evaluate whether the context-enhanced transformer can enhance the performance of sentence extraction tasks in different scenarios, this study employs the training datasets comprising 50, 100, 200, and 500 samples, as constructed in Section 6.1, to train context-enhanced transformer models.

#### (1) The impact of context on the importance of words within sentence

To assess the contribution of the context-enhanced transformer, we revisit the example in Table 2. The detailed information is provided in Table 15. The prediction made by the PLM-BiLSTM-ATT model incorrectly identifies the sentence as a problem sentence. In contrast, both the concatenation model and the context-enhanced transformer provide the correct prediction, aided by the context.

**Table 12. Case of the context-enhanced transformer. The underlined word indicates the method in the target sentence and the method-related content in the context.**

<table>
<tr><th>Sentence category</th><th>Sentence</th><th>Prediction results</th></tr>
<tr><td>Target sentence</td><td><u>Glosser</u> is designed to support reading and learning to read in a foreign language.</td><td rowspan="3">Human annotation:<br>cooccurrence sentence<br>PLM-BiLSTM-ATT:<br>problem sentence<br>Concatenation model:<br>cooccurrence sentence<br>context-enhanced transformer:<br>cooccurrence sentence</td></tr>
<tr><td>Preceding sentence</td><td>None</td></tr>
<tr><td>Following sentence</td><td>There are four language pairs currently supported <u>by Glosser</u>: English-Bulgarian, English-Estonian, English-Hungarian and French-Dutch</td></tr>
</table>

To investigate whether context-enhanced transformers can filter out noise when incorporating essential content into sentences, Figure 6 illustrates the output values of the attention mechanisms in three models: PLM-BiLSTM-ATT, concatenation model, and context-enhanced transformer. Larger output values from the attention mechanism signify the importance of content for problem and method sentence extraction. As depicted in the figure, we observe the following:

**Models combined with context can effectively capture critical information in the context that is relevant to the target sentence.** For example, the attention values for "glosser" and "is designed to support" in both the concatenation model and the context-enhanced transformer are higher than in the PLM-BiLSTM-ATT. In the context, the word "glosser" is present, which also appears in the target sentence. Furthermore, the context includes the word "by," indicating that "glosser" is a method or tool.

**Directly concatenating context with the target sentence introduces noise to the models.** The concatenation model simply combines the context with the target sentence, while the context-enhanced transformer learns relevant content. As shown in Figure 8, the concatenation model assigns a higher attention value to the period (".") compared to PLM-BiLSTM-ATT, whereas the attention value for the period in the context-enhanced transformer is similar to that of PLM-BiLSTM-ATT. Periods are typically considered stop words in sentences, and the concatenation model, which directly combines the context with the target sentence, increases the importance of periods, thereby introducing noise to the models.

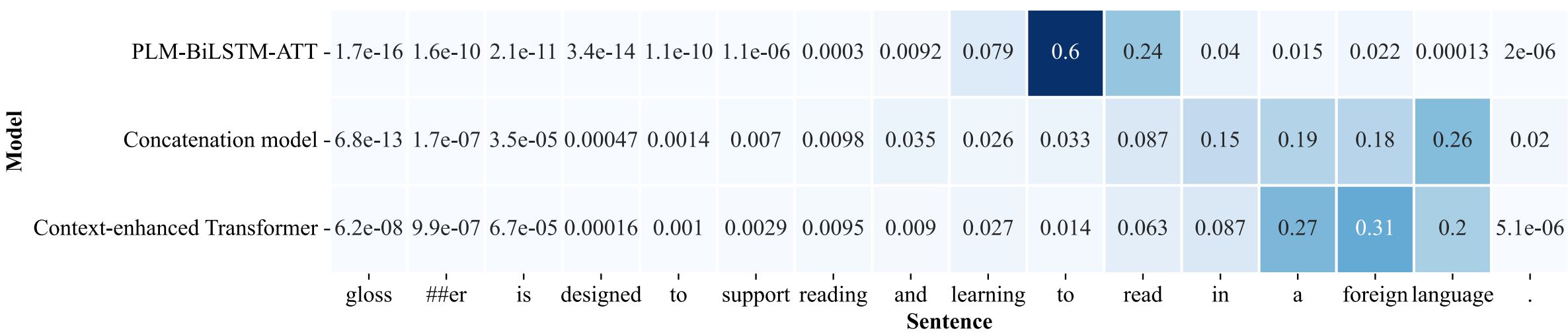

**Figure 8. Heatmap of the attention mechenism output in PLM-BiLSTM-ATT, concatenation model, and context-enhanced transformer.**

**(2) Performace on datasets with different samples**

Figure 9 displays the macro $F_1$ of the PLM-BiLSTM-ATT and the context-enhanced transformer on the abstract and full-text dataset with sample sizes of 50, 100, 200, and 500. As observed in the figure, **when training with datasets of various sample sizes, the macro $F_1$ of the context-enhanced transformer consistently outperforms those of PLM-BILSTM-ATT.** This underscores the effectiveness of the context-enhanced transformer proposed in this paper across different sample sizes.

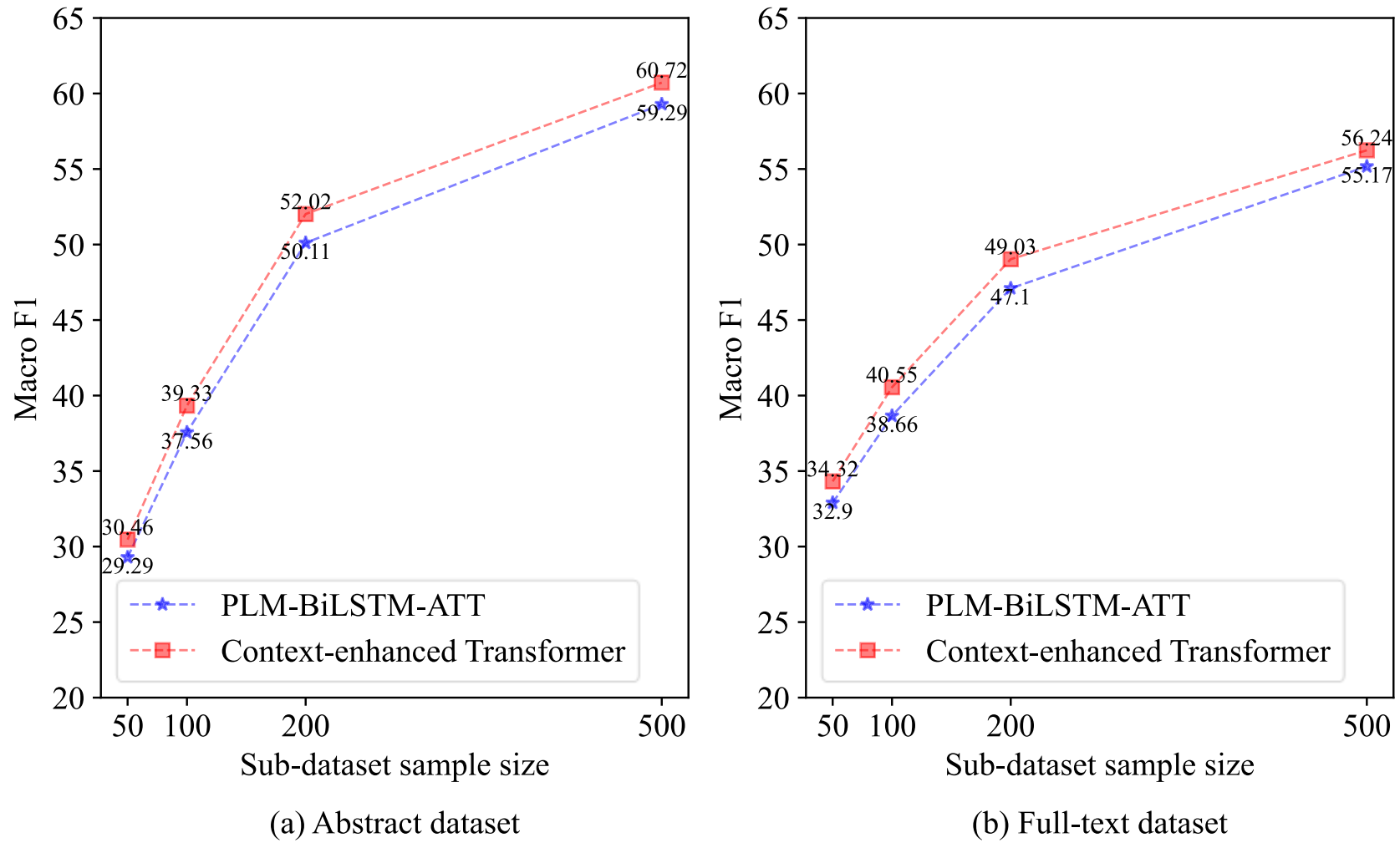

(a) Abstract dataset (b) Full-text dataset

**Fig 9. The macro $F_1$ of PLM-BILSTM-ATT and context-enhanced transformer on abstract and full-text dataset with different samples.**

### 6.3 The error analysis for in-context learning method

In Section 5.4.4, we observed that the macro $F_1$ of the LLM-based ICL method was significantly lower than that of the baseline model in both datasets. To further analyze the effectiveness of the LLM-based ICL method in the task of problem and method sentence extraction, this paper selected four cases and compared the results of the LLM-based ICL method with models trained on 50 samples. As shown in Cases 1, 2, and 3 in Table 13, we observed that the model trained with 50 samples successfully predicted the category of the sentence, while the LLM-based ICL method failed to identify the problems and method content within those sentences. In Case 4 in Table 13, the LLM-based ICL method incorrectly predicted a sentence as a method sentence due to the presence of the word “model”. **These observations indicate that the context learning method based on ChatGPT is not well-suited for the task of problem and method sentence extraction.**

**Table 13. The results of the four cases predicted by models trained with 50 samples and LLM based ICL methods. The bold words indicate the problem and method-related content within the sentences.**

| num | Sentence | Human annotation | 50 samples | LLM based ICL methods |
|---|---|---|---|---|
| 1 | We apply a **decision tree based approach** to **pronoun resolution in spoken dialogue** . | MT | MT | M |
| 2 | **Best-first search** is a common search framework . | M | M | O |
| 3 | Our final task was to **design a system of translation** | T | T | O |
| 4 | Table 4 shows the overall performance of 18 different models for 50 topics | O | O | M |

### 6.4 Time complexity analysis on the proposed method and baseline models

This section compares the time complexity of the proposed Context-enhanced Transformer model against the baseline models. The time complexity is determined by measuring the duration each model takes to process a batch during training and testing, recorded in seconds. As shown in Table 14, first, models incorporating sentence context, including the Sequential model, Concatenation model, and Context-enhanced Transformer, exhibit higher time complexity compared to models like PLM-BiLSTM-ATT and PLM-BiLSTM-TRANS. Second, among the context-based models, the time complexities are comparable. Thus, the Context-enhanced Transformer model is able to improve the performance of sentence classification tasks without increasing the time complexity.

**Table 14. Time complexity on the proposed model and baseline models. The "seconds/one batch (for training)" and "seconds/one batch（for training）" represent the time taken to process each batch during training and testing, respectively.**

| **Model** \ **Time complexity** | seconds/one batch (for training) | seconds/one batch (for testing) |
|---|---|---|
| PLM-BiLSTM-ATT | 0.007 | 0.006 |
| PLM-BiLSTM-TRANS | 0.008 | 0.007 |
| Sequential model | 0.020 | 0.017 |
| Concatenation model | 0.020 | 0.017 |
| Context-enhanced Transformer | 0.020 | 0.017 |

# 7. Conclusion and future work

This paper aims to extract problem and method sentences from scientific papers and addresses four research questions. It has four research questions. First, in response to the first question, this paper conducted experiments on two manually annotated datasets. The experimental results found that FE desensitization-based data augmenter can reduce the model's dependence on FEs, thereby increasing the macro $F_1$ on problem and method sentence extraction tasks by 3.36% and 1.9%, respectively. In response to the second issue, experiments on two manually annotated datasets showed that the context-enhanced transformer can effectively combine context with sentences while reducing the noise in the context, thereby increasing the macro $F_1$ on sentence extraction tasks by 2.83% and 1.52%, respectively. In response to the third question, combining the proposed data augmenter with the context-enhanced transformer can increase the macro $F_1$ on two datasets by 3.71% and 2.67%, respectively. Regarding the fourth question, the results in two datasets showed that LLM based ICL is not suitable for problem sentence and method sentence extraction tasks.

This paper provides the following implications from the above conclusion: (1) The FE desensitization-based data augmenters and the context-enhanced transformer can be transferred to sentence extraction tasks in other fields. (2) The abstract and full-text datasets can serve as valuable resources for future research. (3) The paper underscores the importance of diminishing the model's dependency on FEs, and also emphasize that FEs extracted from the source sentence have a more favorable impact on the model compared to those obtained from open-source datasets. (4) This paper underscores the adverse effects of directly concatenating context with target sentences. These observation can help subsequent researchers in avoiding this practice.

This paper has some limitations. First, this paper aims to reduce the models' dependence on FEs. Since FEs contain essential semantic information, replacing all FEs will negatively impact models. In the future, we plan to use reinforcement learning to select FEs strategically. Second, this paper regards adjacent sentences as the context. However, not only are adjacent sentences, but the entire paragraph may be related to the target sentence. Therefore, in the following research, we can generate a summary of the context and introduce this summary into the target sentence. Third, the field of experimental datasets is limited due to the high labeling cost. In the future, we will conduct experiments in more fields, such as chemistry and physics. Transfer learning methods can be designed to leverage two datasets constructed in this paper.

## Acknowledgments

This work is supported by National Natural Science Foundation of China (Grant No.72074113).

## Ethics declarations

**Conflict of interest**: The authors have no competing interests to declare that are relevant to the content of this article.